\documentclass{article}

\usepackage{microtype}
\usepackage{graphicx}
\usepackage{subfigure}
\usepackage{booktabs} 
\usepackage{hyperref}

\usepackage[accepted]{icml2024}

\usepackage{amsmath}
\usepackage{amssymb}
\usepackage{mathtools}
\usepackage{amsthm}

\usepackage[capitalize,noabbrev]{cleveref}

\theoremstyle{plain}
\newtheorem{theorem}{Theorem}[section]
\newtheorem{proposition}[theorem]{Proposition}

\theoremstyle{definition}
\newtheorem{definition}[theorem]{Definition}

\theoremstyle{remark}

\usepackage{verbatim}

\usepackage{pst-node}
\usepackage{tikz-cd}

\usepackage{graphicx}
\usepackage{wrapfig}
\usepackage{amsthm,amsmath,amssymb}
\usepackage{sidecap, caption}

\icmltitlerunning{Interpreting Equivariant Representations}

\begin{document}

\twocolumn[
    \icmltitle{Interpreting Equivariant Representations}
    \icmlsetsymbol{equal}{*}
    
    \begin{icmlauthorlist}
    \icmlauthor{Andreas Abildtrup Hansen}{xxx}
    \icmlauthor{Anna Calissano}{zzz,yyy}
    \icmlauthor{Aasa Feragen}{xxx}
    \end{icmlauthorlist}
    
    \icmlaffiliation{xxx}{Department of Visual Computing, Technical University of Denmark, Kgs. Lyngby, Denmark}
        \icmlaffiliation{zzz}{INRIA d'Université Côte d'Azur, France}
    \icmlaffiliation{yyy}{Now at: Department of Mathematics, Imperial College London, London, England}
    
    \icmlcorrespondingauthor{Andreas Abildtrup Hansen}{andab@dtu.dk}
    
    \icmlkeywords{Machine Learning, ICML}
    
    \vskip 0.3in
]

\printAffiliationsAndNotice{}  

\begin{abstract}
Latent representations are extensively used for tasks like visualization, interpolation, or feature extraction in deep learning models. This paper demonstrates the importance of considering the inductive bias imposed by an equivariant model when using latent representations as neglecting these biases can lead to decreased performance in downstream tasks. We propose principles for choosing invariant projections of latent representations and show their effectiveness in two examples: A permutation equivariant variational auto-encoder for molecular graph generation, where an invariant projection can be designed to maintain information without loss, and for a rotation-equivariant representation in image classification, where random invariant projections proves to retain a high degree of information. In both cases, the analysis of invariant latent representations proves superior to their equivariant counterparts. Finally, we illustrate that the phenomena documented here for equivariant neural networks have counterparts in standard neural networks where invariance is encouraged via augmentation.
\end{abstract}

\section{Introduction}
\begin{figure}
    \centering
    \includegraphics[width=0.32\textwidth]{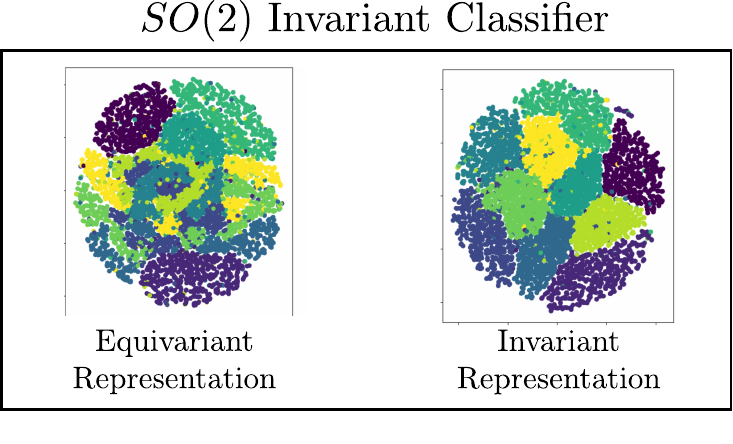}
        \vspace{-0.2cm}
    \caption{Visualizing rotated MNIST images via their equivariant representation (left) hides structure that is apparent in an invariant representation of the same latent codes (right).}
    \label{fig:intr_visualization_so2}
\end{figure}

Latent representations are used extensively in the interpretation and design of deep learning models. The latent spaces of VAEs~\citep{kingma2019introduction} and masked autoencoders~\citep{he2022masked}, learned from large amounts of unlabeled data unsupervised or self-supervised, are prominent examples of powerful latent representations. These representations are used for e.g.~molecular-, chemical- or protein discovery~\citep{detlefsen2022learning}; image generation, synthesis~\citep{goodfellow2020generative} and segmentation~\citep{kirillov2023segment}; semantic interpolation \citep{Berthelot2018-hh}; interpretability using visualization tools such as T-SNE~\citep{tsne}, UMAP~\citep{mcinnes2018umap} or PCA on latent embeddings; or counterfactual explainability~\citep{papernot2018deep}.

However, the analysis of such latent representations is no trivial task, and may yield misguided conclusions if done naively. Usually, deep neural networks are designed with specific geometric inductive biases in mind~\cite{Bronstein2021-fi}; an inductive bias can be imposed on a deep learning model by requiring the model to be invariant/equivariant to certain transformations of the data~\citep{pmlr-v48-cohenc16,puny2022frame} represented by a group of symmetries. A consequence of incorporating such a property in the modelling procedure is to make sure that datapoints which are \textit{similar} are be mapped to \textit{similar} outputs. This observation has consequences for any representation we might want to learn, as we will in general aim for any learned representation to be invariant to the transformation in question, while maximizing expressive power~\cite{wang2024rethinking}. These invariant representations can be obtained in many ways using for instance data augmentation or an inherently invariant model~\cite{Kwon2023-gv} or by quotienting out nuisance parameters~\cite{Williams2021-ow}, however, in the case of autoencoders -- in this paper we use VAEs as a running example -- it is natural design these to be equivariant, and thus the question of interest becomes: How to we obtain an invariant representation using an equivariant model?

\paragraph{We contribute:} An empirical demonstration of how a naive interpretation of equivariant latent spaces can lead to incorrect conclusions about data and decreased performance of derived models. We mathematically explain how analysis of equivariant latent representations needs to take the group action into account. We provide explicit tools for doing so using invariant projections of latent spaces. We evaluate the effect of the suggested tools via widely encountered group actions on two widely used model classes: 1) A permutation equivariant variational autoencoder (VAE) representing molecular graphs acted on by node permutations, where we obtain isometric invariant representations of the data, and 2) an equivariant representations of a rotation-invariant image classifier, where we showcase random invariant projections as a general and efficient tool for providing expressive invariant representations. Finally, we show that the ambiguity of equivariant latent representation also extends to standard deep learning models, where invariance is encouraged via augmentation. An empirical example shows that even for such models, latent representations display behavior similar to that observed in equivariant latent spaces. Thus, while these ambiguities might be known and avoided by experienced developers of equivariant models, pointing them out and providing tools to manage them is important to all users that encode implicit biases via equivariance or augmentation.

\paragraph{Invariance and equivariance:} A geometric inductive bias can be imposed on deep learning models by requiring the model to be either invariant or equivariant to certain transformations of the data, usually represented by a group of symmetries~\citep{pmlr-v48-cohenc16,puny2022frame}. In invariant models, the prediction is unchanged when group transformations act on the model's input. Mathematically, we formalize this as follows: If $h \colon \mathcal{X} \to \mathcal{Y}$ is a neural network with a group $G$ acting on the input space $\mathcal{X}$, then $h$ is $G$-\emph{invariant} if $h(g \cdot x) = h(x)$ for all $g \in G$. In equivariant models, the prediction has an analogous action of the group $G$ on the target space $\mathcal{Y}$, whose transformations are aligned with those acting on the inputs  -- as is the case, for instance, with equivariant autoencoders. Then $h$ is $G$-\emph{equivariant} if $h(g \cdot x) = g \cdot h(x)$, aligning the predictions with their inputs according to the group action $G$.

\paragraph{Equivariant representations:} Both invariant and equivariant networks are commonly designed with latent feature representations $\mathcal{Z}$, namely $h \colon \mathcal{X} \xrightarrow{f} \mathcal{Z} \xrightarrow{k} \mathcal{Y}$, where the latent space $\mathcal{Z}$ has an action of $G$, and the feature embedding $f$ is $G$-equivariant. We refer to the latent representation $\mathcal{Z}$ as an \emph{equivariant representation}. Equivariant latent representations come with the caveat that any latent code $z = f(x) \in \mathcal{Z}$ will be equivalent to the, often different, representation $g \cdot z = g \cdot f(x) = f(g \cdot x)$ of the transformed, but equivalent, input $g \cdot x$. Thus, each input $x$ will be represented by the entire set $G \cdot z = \{g \cdot z | g \in G \}$ of latent vectors acted on by the group $G$. Different choices of $g$, which are typically not made by the user but rather determined implicitly by the data collection, can lead to widely different latent embeddings $g \cdot z$, see Fig.~\ref{fig:intr_visualization_so2}. 

\section{Equivariant Models Enable Implicit Quotient Learning}
The modelling-assumption that two elements $x_1$ and $x_2$ are equivalent, if there exists a group element~$g$ transforming one into the other is encapsulated by the \emph{quotient space} $\mathcal{X}/G$, which is the set consisting of all orbits $G \cdot x =[x] = \{g \cdot x | g \in G\}$~\citep{bredon1972introduction}. While quotient spaces have been used extensively in classical approaches to statistics and machine learning with geometric priors, they often come with non-Euclidean structure and singularities, severely inhibiting the availability of tools for statistics, optimization, and learning~\citep{feragen2020statistics, mardia1989statistical,kolaczyk2020averages,severn2022manifold,calissano2024populations}. Equivariant models implicitly encode the structure of quotient spaces, while, from the viewpoint of implementation and optimization, Euclidean tools are available with all the computational advantages they may offer.

In fact, the modeling choice of picking an equivariant feature embedding $f$ implicitly induces a feature embedding $f'$ defined on the quotient spaces of the input space $\mathcal{X}$ and the latent space $\mathcal{Z}$:
\[
\begin{tikzcd}
\mathcal{X} \arrow{r}{f} \arrow[swap]{d}{\pi} & \mathcal{Z} \arrow[swap]{d}{\pi}\\%
\mathcal{X}/G_{\mathcal{X}} \arrow[dashed]{r}{f'} & \mathcal{Z}/G_{\mathcal{Z}} \\
\end{tikzcd}
\]
Where $\pi$ denotes the canonical projection. The induced feature embedding $f'$ can be defined via representatives $x$ from each orbit. That is,
\[
f'([x]) = [f(x)] \in \mathcal{Z}/G
\]
for every orbit $[x] \in \mathcal{X}/G$. The quotient space $\mathcal{Z}/G$ comes with a \emph{quotient metric} defining distances between orbits:
\[
d([z_1], [z_2]) = \min_{g_1, g_2 \in G} \|g_1 \cdot z_1 - g_2 \cdot z_2\| = \min_{g \in G} \|z_1 - g \cdot z_2\|.
\]
As the quotient spaces $\mathcal{X}/G$ and $\mathcal{Z}/G$ can exhibit strongly non-Euclidean geometry~\cite{calissano2024populations}, it is usually not feasible to fit a model mapping between these two spaces directly. However, by defining $f$ to be equivariant we do effectively learn a map $f'$ between the quotient spaces, without having to deal with the issues that the non-Euclidean geometry of the quotient space causes.

\section{Equivariant Latent Representations are Ambiguous}
\label{sec:background}
We have previously stated that, in general, we prefer invariant representations to equivariant representations, but why is that? This is since the interpretation and utilization of an equivariant latent representation is non-trivial, as any two points $x_1, x_2 \in \mathcal{X}$, which are equal up to some group element $g \in G$, i.e.~$x_1 = g \cdot x_2$, will almost certainly \textit{not} map to the same latent representation -- they may in fact map to latent representations $z_1 = f(x_1)$ and $z_2 = f(x_2) = g \cdot f(x_1)$ which are far apart in the latent space. If standard latent-space analysis methods are applied out of the box on such equivariant representations, it can create problems with downstream analysis, which typically relies on similar data points having nearby latent representations. An obvious example of this would be clustering based on the latent representations which often would rely on the pairwise Euclidean distances between the latent codes in the creation of the clusters.

\begin{figure}
    \centering
    \includegraphics[width=\linewidth]{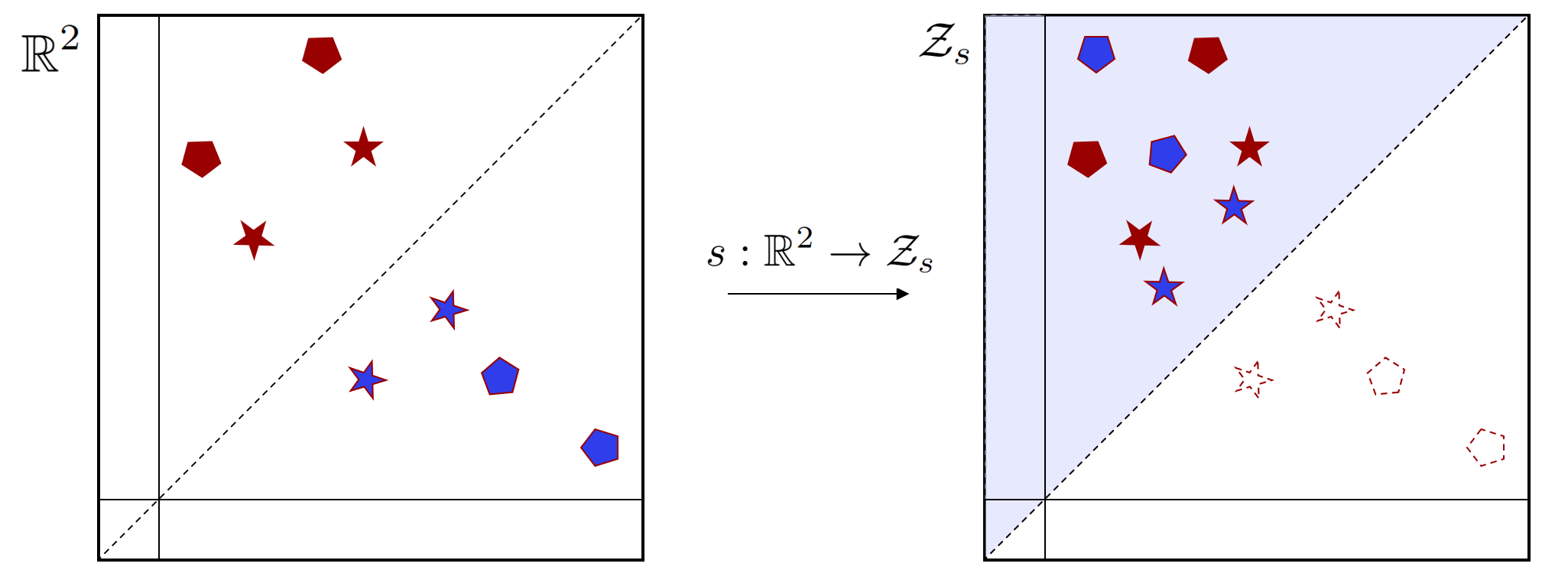}
    \caption{An illustration of the effect of applying sorting to $\mathbb{R}^2$. In this case, we see, that $\mathbb{R}^2$ is mapped to $\mathcal{Z}_s = \{(x,y) \in \mathbb{R}^2 | x \leq y\}$ (i.e. the blue area). We see, that in the $S_2$ equivariant representation depicted on the left-hand side the similar objects (e.g. the pentagons) are not guaranteed to be close. After having applied an invariant map (sorting), we see, that this shortcoming is taken care of.} 
    \label{fig:R2modS2}
\end{figure}

In Figure \ref{fig:R2modS2} we illustrate the problem on a simple example: the equivariant latent representation $\mathcal{Z}$ is analyzed using arbitrary equivariant representatives $g_1 \cdot z_1$ and $g_2 \cdot z_2$ of the orbits $[z_1]$ and $[z_2]$, the relative distances between the latent codes $g_1 \cdot z_1$ and $g_2 \cdot z_2$, which would typically be used for downstream analysis, are not well defined, as they depend on the group elements $g_1$ and $g_2$ -- and may be very different from the relative quotient distance $d([z_1], [z_2])$ within $\mathcal{Z}/G$. Thus, it is key that we extend our modelling assumption, about which data symmetries are present, to our representation before conducting any down-stream analysis of the data. In practice this is done by considering an invariant representation.

\section{Invariant Analysis of Equivariant Representations}
\label{sec:invariant_analysis}

As outlined in the previous sections, the relative distance between latent equivariant representations is ill-defined, due to the multiple equivalent representations of data. In this section, we discuss how invariant projections of the equivariant latent representation can be used to obtain invariant representations that give unambiguous latent embeddings.

As any equivariant function composed with an invariant function will be invariant, we can obtain invariant representations of our latent features by passing them through an invariant function. That is, our objective is to find an invariant map $s \colon \mathcal{Z} \to \mathcal{Z}_s$, and use this to extract invariant latent features. An obvious invariant projection is the quotient projection $\pi \colon \mathcal{Z} \to \mathcal{Z}/G$, but, as discussed above, the quotient $\mathcal{Z}/G$ comes with severe limitations for analysis. Instead, we seek an invariant feature representation $s$ where the invariant latent space $\mathcal{Z}_s$ is Euclidean. Note that any such $s$ can necessarily be written as a composition of $\pi$ with another mapping $s'$ as illustrated by the following proposition:

\begin{proposition}
\label{prop:invariant_surjective}
Let $s: \mathcal{Z} \to \mathcal{Z}_s$ be an invariant, surjective function. Then $s$ induces a surjective function $s': \mathcal{Z}/G \to \mathcal{Z}_s$ as
    \begin{equation}
        s'([z]) = s(z) \qquad \forall [z] \in \mathcal{Z}/G.
    \end{equation}
That is, the following diagram commutes:
\[ \begin{tikzcd}
\mathcal{Z} \arrow{r}{s} \arrow[swap]{d}{\pi} & \mathcal{Z}_s \\%
\mathcal{Z}/G \arrow{ru}{s'}
\end{tikzcd}
\]
We will refer to elements of $\mathcal{Z}$ as equivariant representations and to elements of $\mathcal{Z}_s$ as invariant representations.
\end{proposition}
A proof of this proposition can be found in Appendix \ref{proof:inv_sur}. As the map $s$ is chosen post hoc, a main challenge is how to choose $s$ and $\mathcal{Z}_s$ to retain the signal from the quotient space $\mathcal{Z}/G$ while simplifying the analysis. Choosing $s(z) = 0$ is obviously invariant, but will destroy any signal in $\mathcal{Z}$. Prop.~\ref{prop:invariant_surjective} highlights one of the problems of picking an arbitrary invariant map $s$, as we will in general only be guaranteed that $\mathcal{Z}_s$ is more coarse than $\mathcal{Z}/G$ and that the neighbourhood structure in $\mathcal{Z}_s$ may be substantially different from the quotient $\mathcal{Z}/G$. Ideally, we would pick a map $s: \mathcal{Z} \to \mathcal{Z}_s$, which in turn induces an \textit{isometric embedding} $s': \mathcal{Z}/G \to \mathcal{Z}_s$. However, one might consider what to do in cases, where such an embedding does not exist. In these cases we might relax the requirement that distances between points should be preserved and only require that $s'$ is a homeomorphism. If we can ensure that this is the case, then we will effectively have preserved the neighborhood structure of the quotient space, which is in itself a very powerful guarantee. Luckily, classical point set topology, e.g.~\cite{munkres2000topology}, provides us with the recipe for ensuring that this is true. For the sake of clarity we first define:

\begin{definition}
\label{def:isometric_embedding}
Let $X$ and $Y$ be metric spaces with metrics $d_X$ and $d_Y$. Assume that the map $f: X \to Y$ is a \textit{homeomorphism}. Then $f$ is an embedding. If $f$ has the property that:
\begin{equation}
    d_X(x_1, x_2) = d_Y(f(x_1), f(x_2)) \text{ for all } x_1, x_2 \in X
\end{equation}
Then $f$ is an \textit{isometric embedding}.
\end{definition}

\begin{figure*}
    \centering
    \includegraphics[width=0.9\textwidth]{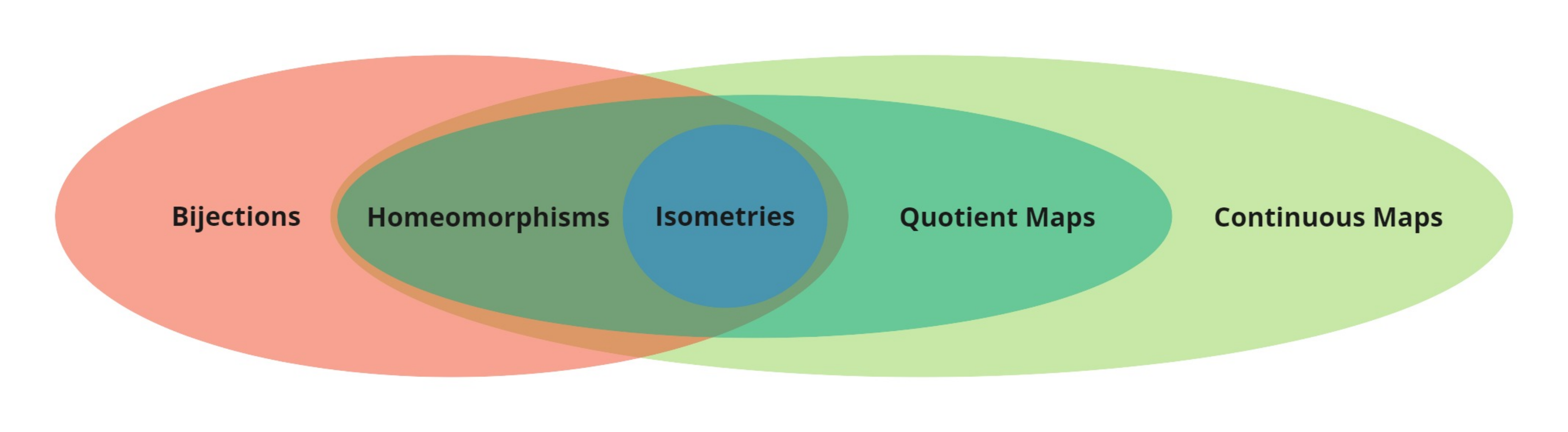}
    \caption{An illustration of possible properties of the induced map $s': \mathcal{Z}/G \to \mathcal{Z}_s$ to prioritize when choosing an invariant projection $s: \mathcal{Z} \to \mathcal{Z}_s$. Each category of mappings preserves a different amount of structure of the latent space.} 
    \label{fig:venn}
\end{figure*}

\begin{definition}
    Let $\mathcal{Z}$ and $\mathcal{Z}_s$ be topological spaces, and let $f: \mathcal{Z} \to \mathcal{Z}_s$ be a surjective map. Then $f$ is said to be a \textit{quotient map} provided $U \subseteq \mathcal{Z}_s$ is open if and only if $f^{-1}(U) \subseteq \mathcal{Z}$ is open.
\end{definition}
\begin{proposition}
\label{prop:cont-quotient-homeo}
    Let $s: \mathcal{Z} \to \mathcal{Z}_s$ be defined as in proposition $\ref{prop:invariant_surjective}$. Then the induced map $s'$ is continuous if and only if $s$ is continuous; $s'$ is a quotient map if and only if $s$ is a quotient map; $s'$ is a homeomorphism if and only if $s'$ is a bijective quotient map. 
\end{proposition}
A proof of Proposition \ref{prop:cont-quotient-homeo} can be found in Appendix \ref{proof:cont-quotient-homeo}. We can now use Proposition $\ref{prop:cont-quotient-homeo}$ as a guideline for how to select an invariant map $s$ that preserves the topology of the quotient space in our invariant representation. An illustration of the possible choices can be seen in Fig. \ref{fig:venn}, where an invariant isometry would yield an invariant latent representation most faithful to the original. Note that the hierarchy above illustrates what to take into account when picking an invariant map which respects the topology of the quotient space to as large a degree as possible, however, if there exists no embedding (i.e. homeomorphism) of the quotient space, then one might choose to sacrifice neighbourhood information in favour of picking a bijective map, as in general continuous- and quotient maps $s(z) \in \mathcal{Z}_s$ does not uniquely identify the orbits of an element $z \in \mathcal{Z}$.

\subsection{Retrieving an Isometric Cross Section: Latent Graph Representation}
\label{sec:latent_graph}
In a special case visited below, where $\mathcal{Z}_s$ and $\mathcal{Z}/G$ are isometric and $\mathcal{Z}_s \subset \mathcal{Z}$, we will speak of~$\mathcal{Z}_s$ as being an isometric \textit{cross section}. In cases where such a cross section exists, we argue, that the equivariant representation of the latent features should be mapped onto the cross section prior to any subsequent analysis, as this will be analogous to considering an isometric embedding.

We consider the special case of a permutation equivariant model $h~\colon~\mathcal{X}~\xrightarrow{f}~\mathcal{Z}~\xrightarrow{k}~\mathcal{X}$, where the group $G$ is the symmetric group $S_n$ of permutations on $n$ elements, and the latent representation is given by $\mathcal{Z} \subseteq \mathbb{R}^{n}$. For this particular case, we show, that we can choose an invariant map $s$ which induces an isometric cross section $\mathcal{Z}/G \to \mathcal{Z}_s \subset \mathcal{Z}$. Let $s \colon \mathcal{Z} \to \mathcal{Z}_s $ be defined as
\begin{equation}
    s(z) = \sigma_z \cdot z,
\end{equation}
where $\sigma_z \in S_n$ is a permutation which ensures that 
\begin{equation}
    z_{\sigma_z(1)} \leq z_{\sigma_z(2)} \leq ... \leq z_{\sigma_z(n)}.
\end{equation}
In other words, $\sigma_z$ is the permutation that sorts the coordinates of $z$ in ascending order. This permutation clearly exists, since all sequences can indeed be sorted. While the sorting permuation $\sigma_z$ need not be unique, the sorted sequence will always be unambiguous.

The resulting map $s$ is clearly invariant, since $\sigma z$ and $z$ will have the same form when sorted for all $\sigma \in S_n$. Also, if we let $\mathcal{Z}_s = \{ s(z) \mid z \in \mathcal{Z} \}$, then $s$ is surjective by definition. These observations, combined with Proposition \ref{prop:invariant_surjective}, allow us to show the following results implying that $s$ does indeed induce an isometric cross section $s'\colon \mathcal{Z}/S_n \to \mathcal{Z}_s$:

\begin{proposition}
Let $s\colon \mathcal{Z} \to \mathcal{Z}_s$ be the sorting function described above. Furthermore, equip $\mathcal{Z}$ and $\mathcal{Z}_s$ with the Euclidean metric, and $\mathcal{Z}/S_n$ with the quotient metric. Then the induced $s'\colon \mathcal{Z}/S_n \to \mathcal{Z}_s$ defined as in Proposition \ref{prop:invariant_surjective} is a bijection and an isometry. Finally $\mathcal{Z}_s$ is a convex cone.
\label{prop:graph-bijective-iso}
\end{proposition}

A proof of Proposition \ref{proof:prop:graph-bijective-iso} is included in Appendix \ref{app:proofs}. The realization that $\mathcal{Z}_s$ is a convex is important, because it makes linear interpolation between elements of $\mathcal{Z}_s$ meaningful: Any point on the the line will be contained in the $\mathcal{Z}_s$ as well. 

\subsection{A Choice of Continuous Map: Random Invariant Linear Projections}

We cannot in general design an isometric cross section $s'\colon \mathcal{Z}/G \to \mathcal{Z}_s \subset \mathcal{Z}$, and for this more general situation we propose random invariant projections as a generic tool. Random projections are a well known alternative to trained dimensionality reduction techniques~\citep{candes2006near}, and can be easily adapted to equivariant latent representations by using random invariant projections.

Random projections are often available: ~\cite{Ma2018-tq, Maron2019-zr} propose a basis for all permutation invariant linear maps, and~\citep{Cesa2022-to} describe how $E(n)$-equivariant and invariant linear maps can be constructed from a basis. Initializing these layers at random, we obtain an analogy to the random projections known from classical statistics~\citep{candes2006near}.

As we cannot generally invert invariant random projections, they do not allow interpolation-based analysis. However, they are still highly valuable for visualization and interpretation, as well as building new models and analyses directly from the invariant latent representation.

\section{Relation to Existing Invariant-, Quotient- and Equivariant Latent Spaces}

Having presented our methods and notation, we can now explain in detail how they relate to recent related work. While the utility and interpretability of equivariant latent spaces is, to the best of our knowledge, unexplored in the past, the autoencoder literature explores latent space design. \cite{mehr2018manifold} design a quotient autoencoder for 3D shapes whose latents reside on the quotient space $\mathcal{X}/G$. Here, $\mathcal{X}$ parametrizes 3D shape and $G$ is the group of rotations or non-rigid deformations. As the latent space is $G$-invariant, shape alignment and interpolation are greatly simplified. However, as discussed above, quotient spaces are often non-Euclidean, greatly hindering their applicability.

Graph autoencoders commonly use a permutation \textit{invariant} encoder. This can be achieved by using permutation equivariant layers (e.g. graph-convolutions) eventually followed by a permutation invariant layer~\cite{Winter_undated-pz}. As the composition of equivariant and invariant functions is invariant, this ensures that the latent representation is indeed invariant to permutation of the input nodes. This was the strategy followed e.g.~in early graph VAE models~\cite{simonovsky2018graphvae,Vignac2021-lp,Rigoni2020-hg,Liu2018-rj}, where an expensive graph alignment step was needed in order to train the model. To counteract this, newer models~\cite{hy2023multiresolution} replace the invariant latent space with equivariant ones, similar to those studied in this paper. Similarly to the early VAE models, \cite{Winter2022-pl} construct encoder-decoder architectures; here, however, the needed alignment of outputs with inputs is learned rather than optimized.

It is important to note that mathematically -- ignoring implementation challenges -- \textbf{all the above approaches are essentially equivalent}. If, as above, $h \colon \mathcal{X} \xrightarrow{f} \mathcal{Z} \xrightarrow{g} \mathcal{Y}$ is a predictor with equivariant latent feature embedding $f$, then the quotient map $\pi \colon \mathcal{Z} \to \mathcal{Z}/G$ composes with $f$ to form a quotient latent feature embedding $f_{quot} = \pi \circ f \colon \mathcal{X} \to \mathcal{Z}/G$ as was done in~\cite{mehr2018manifold}. \cite{sannai2021improved} discusses the generalization bounds for invariant and equivariant networks using the quotient space $\mathcal{Z}/G$. The equivariant representation $\mathcal{Z}$ and the quotient latent representation $\mathcal{Z}/G$ carry \emph{exactly} the same information, only with their own individual caveats -- the quotient representation $\mathcal{Z}/G$ will often be non-Euclidean and cumbersome to work with, whereas -- as we have seen -- the equivariant representation $\mathcal{Z}$ does not have well defined representatives for each data point.
\[ 
\begin{tikzcd}
\mathcal{X} \arrow{r}{f} \arrow{dr}[below]{f_{quot}}  &\mathcal{Z} \arrow{r}{g} \arrow[swap]{d}{\pi} & \mathcal{Y} \\%
 & \mathcal{Z}/G \arrow{ru}{} \arrow{r}{f_{proj}} & \mathcal{Z_{\textit{inv}}} \arrow{u}
\end{tikzcd}
\]
\begin{figure*}[t]
        \centering
        \includegraphics[width=\textwidth]{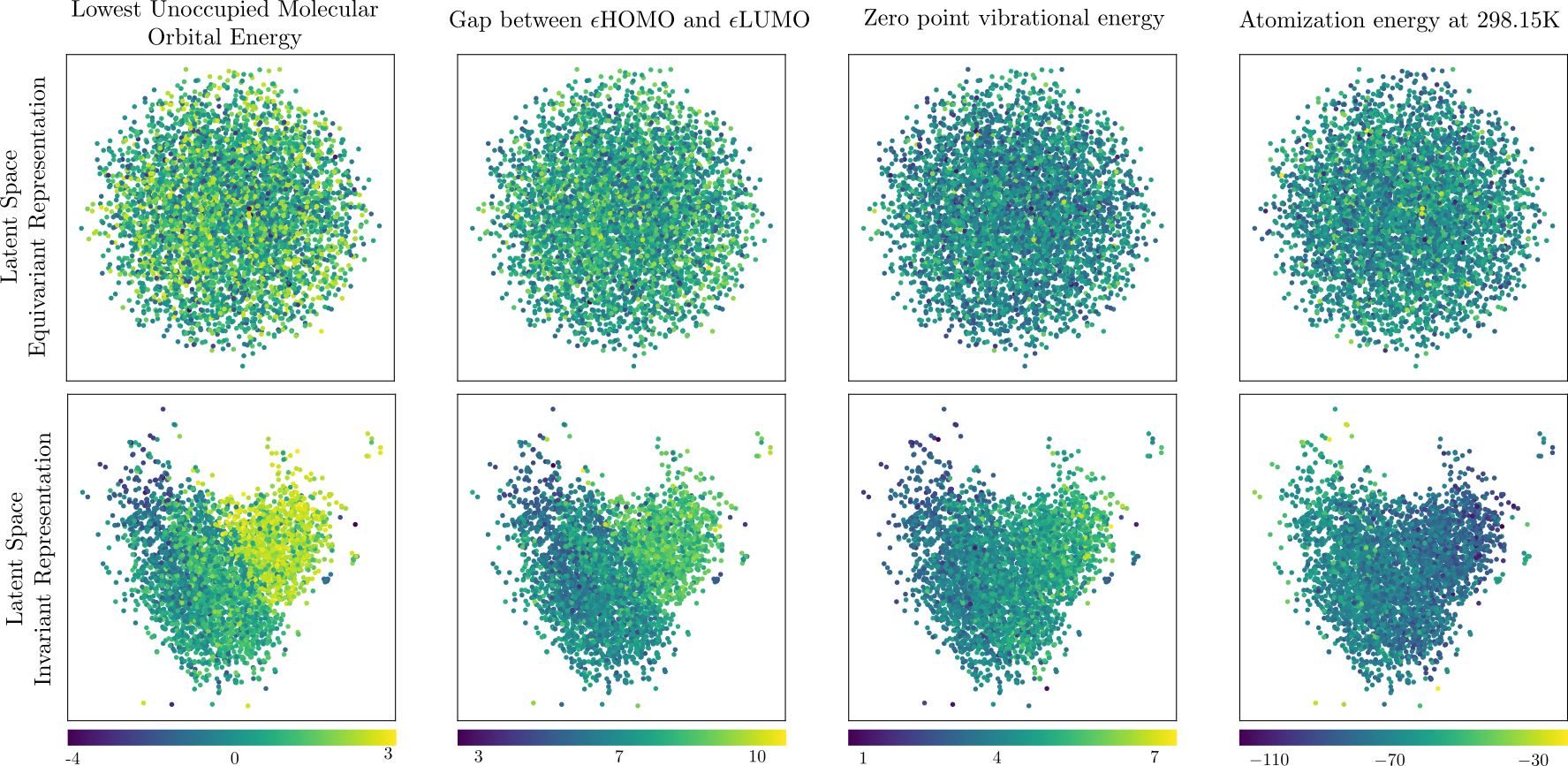}
        \caption{The first two principal components of the QM9 training dataset for the equivariant (top) and invariant representation (bottom) of the latent space. Each column illustrates a specific  molecular property.}
        \label{fig:ml_prop}
\end{figure*}
More generally, any invariant latent feature embedding $f_{inv} \colon \mathcal{X} \to \mathcal{Z}_{inv}$ which carries enough information to enable decoding back onto $\mathcal{X}$, as in~\citep{Winter2022-pl}, can necessarily be written as a composition $f_{proj} \circ f_{quot}$ of a quotient latent feature embedding $f_{quot} \colon \mathcal{X} \to \mathcal{Z}/G$ for some latent space $\mathcal{Z}$, and a projection-like map $f_{proj} \colon \mathcal{Z}/G \to \mathcal{Z}_{inv}$. Thus, any differences in performance as observed in the experiments of~\cite{Winter2022-pl}, are caused by implementation choices rather than differences in the underlying mathematical model -- invariant, quotient and equivariant representations are, mathematically, able to carry the same information. We argue that when taking care to respect the group action when utilizing and interpreting the equivariant latent representation $\mathcal{Z}$, there is no good reason to avoid it.

Another existing approach to obtaining invariant and equivariant maps is using fundamental domains~\cite{aslan2023group}. When isometric cross sections, they form fundamental domains. However, in general, one cannot find isometric cross sections -- indeed, the quotient space can exhibit strongly non-Euclidean geometry~\cite{calissano2024populations}. This explains, in part, why the quotient itself is often not an efficient invariant representation. This also indicates that making an isometric, or even near-isometric, mapping to the representation space mathematically impossible. Random invariant projections, on the other hand, are in our cases designed as continuous mappings from the equivariant feature space and, therefore, preserve some local geometric structure.

\section{Experiments}
\label{sec:exp}
\subsection{Isometric Invariant Representations via Equivariant Grap VAE}

We consider a permutation equivariant graph variational autoencoder (VAE) $h \colon \mathcal{X} \to \mathcal{Z} \to \mathcal{X}$ trained for molecule generation, and evaluate the utility of its latent codes for downstream tasks. A graph $(V, E)\in \mathcal{X}$ consists of a set $V$ of (at most) $n$ nodes with an $n \times d_A$ node feature matrix and an $n \times n \times d_E$ edge feature tensor $E$, where $n$ denotes the number of nodes. A permutation $g \in S_n$ acts on the graph $(V, E)$ through its associated permutation matrix $P_g$:
\begin{equation}
    g(V, E) = (P_g V, P_g E P_g^T).
\end{equation}
The latent space of the VAE is designed as $\mathcal{Z} = \mathbb{R}^{n}$, and we let $s \colon \mathcal{Z} \to \mathcal{Z}_s$ be the invariant isometry defined by sorting function as defined in section \ref{sec:latent_graph}.

\paragraph{Dataset:}
The QM9 dataset~\cite{Ramakrishnan2014-de, Ruddigkeit2012-kh} consists of approx.~130.000 stable, small molecules, using 80\%/10\%/10\%  for training/validation/testing. Each molecule is represented by at most 9 heavy atoms, their bindings (edges), and selected molecular properties. As our interest is in the permutation equivariant representation, we simplify the graphs to contain only atom-type (node features) and binding-type (edge features). Each graphs is padded with not-a-node and not-an-edge features to obtain the same number of nodes.

\paragraph{Model:}
Our permutation equivariant VAE is based on linear equivariant layers as derived in~\cite{maron2018invariant}, combined with entry-wise nonlinearities, which were used in both the encoder and decoder. A comprehensive description of the model architecture can be found in Appendix \ref{app:architecture-vae}.

\paragraph{Visualisation:}
VAEs are often used to visualize disentangled latent representations~\citep{Mathieu2019-pi, Mitton2021-mz}. Here we show how, when working with end-to-end permutation equivariant variational autoencoders, the obtained representations may by deceiving.

\begin{figure*}
    \centering
    \includegraphics[width=0.9\textwidth]{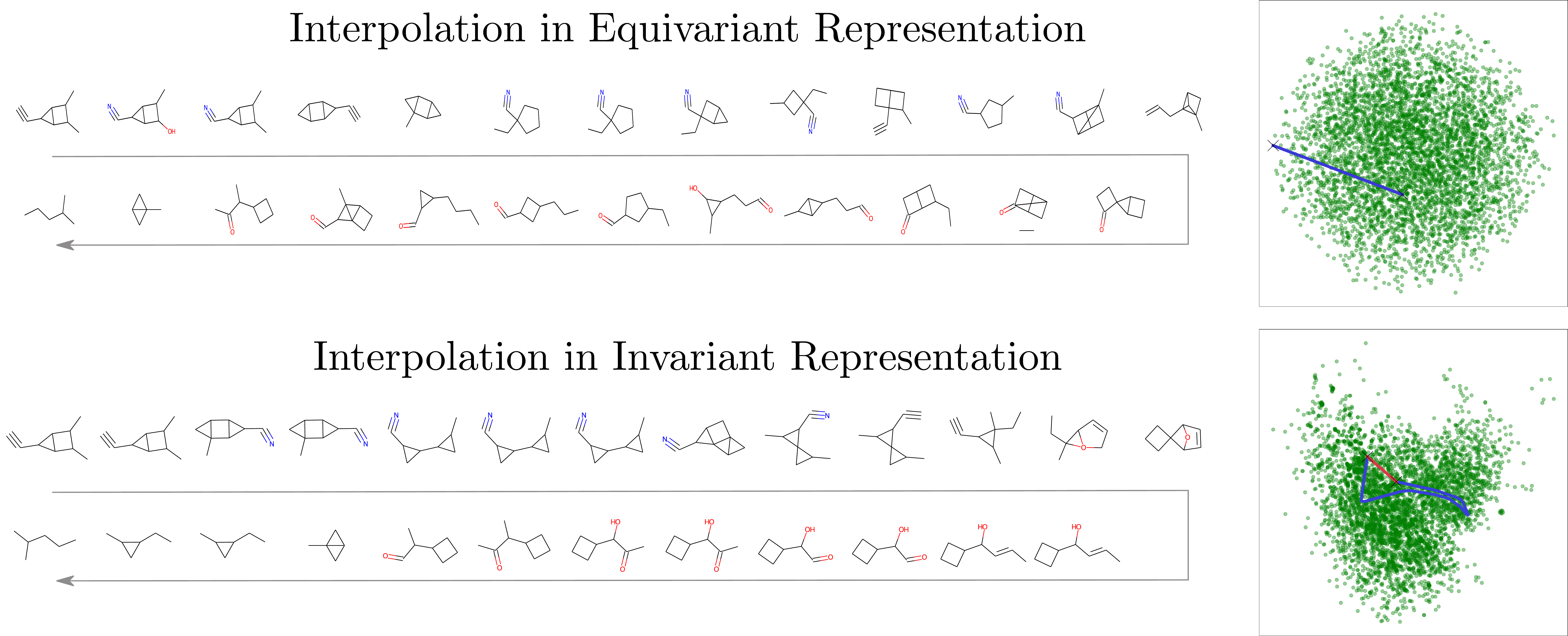}
    \caption{Molecules generated from interpolating between two molecules using the \textit{equivariant} (top) and  \textit{invariant}  (bottom) representations. Note that while the molecules decoded from $z_2$ and $s(z_2)$ differ in their embedding, they are equal up to permutation. Left: Molecules sampled along the two interpolations. Right: Interpolation in the latent space visualized via the first two principal components. In the equivariant representation, we visualize a straight blue line between $z_1$ and $z_2$. In the invariant representation, we visualize the linear interpolation between $s(z_1)$ and $s(z_2)$ (red), and the equivariant linear interpolation between $z_1$ and $z_2$ subsequently mapped to $\mathcal{Z}_s$ (blue).}
    \label{fig:mols_along_curve}
\end{figure*}

The upper row of Fig.~\ref{fig:ml_prop} shows the first two principal components of the equivariant latent representation\footnote{Note that the equivariant representations were obtained by randomly permuting the input nodes to remove any implicit ordering which may have been implied by the structure of the dataset.} $\mathcal{Z}$ of the QM9 test set, with molecular properties encoded by color. Inspecting the equivariant latent representations in the top row incorrectly suggests no apparent structure in the data, as molecules with similar molecular properties are by no means close in the latent space. On the other hand, when inspecting the isometric invariant representation $\mathcal{Z}_s$ plotted in the bottom row, a different picture emerges.  Here, we find a clear pattern between latent representation and molecular properties, indicating that the model does indeed pick up on important structure in the data. In other words, looking at \emph{the same} latent representation using either the equivariant representation, or its invariant projection, leads to very different conclusions about both the data and the model.

\paragraph{Latent Space Interpolation.}
An autoencoder enables straightforward interpolation between molecules: Given two molecular graphs $\mathcal{G}_1, \mathcal{G}_2 \in \mathcal{X}$, we simply linearly interpolate between their respective latent representations and pass the resulting latent representation to the decoder, i.e.
\begin{equation}
    \mathcal{G}_\alpha = g(\alpha f(\mathcal{G}_1) + (1-\alpha)f(\mathcal{G}_2)), \quad \alpha \in [0,1]
\end{equation}
Here, we demonstrate how linear interpolation between latent codes $z_1$ and $z_2$ in the equivariant latent space $\mathcal{Z}$ may yield unstable decoded molecules, while linear interpolation between $s(z_1)$ and $s(z_2)$ in the isometric, invariant representation $\mathcal{Z}_2$ can remedy this issue. Note that linear interpolation between $s(z_1)$ and $s(z_2)$ is indeed meaningful, as we saw in section \ref{sec:invariant_analysis} that $\mathcal{Z}_s$ is convex.

\begin{figure*}[t]
    \centering
    \includegraphics[width=0.65\textwidth]{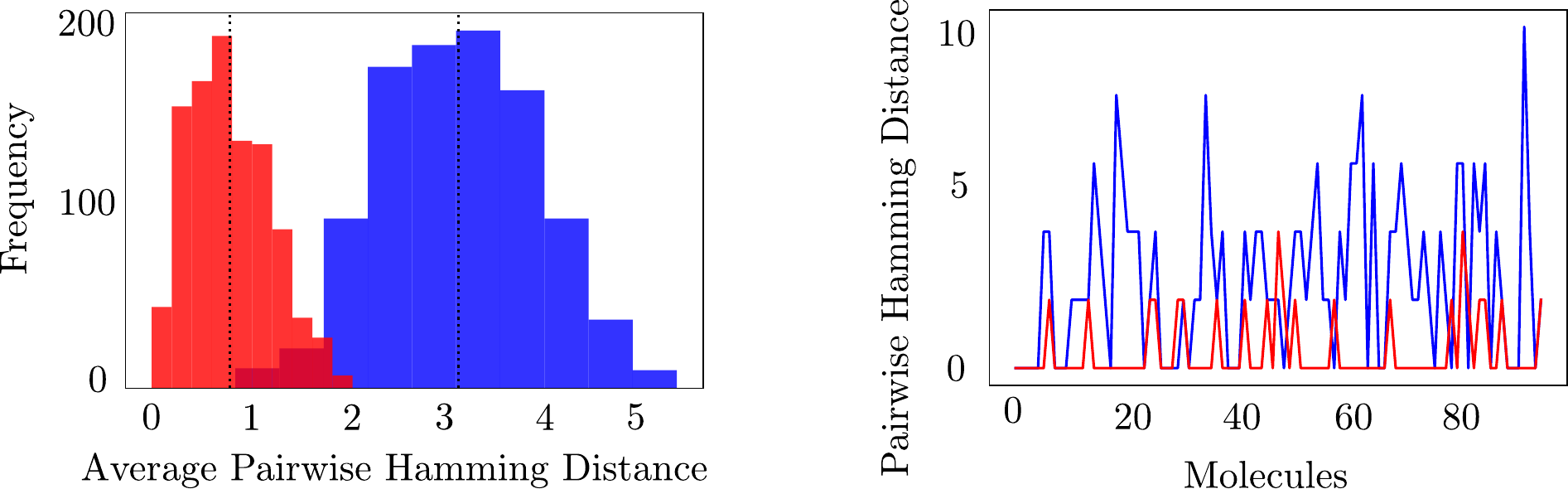}
    \caption{We measure pairwise Hamming distances between consecutive observation pairs sampled along linear interpolations. Right: Histograms over the average pairwise distance for $10000$ interpolations in the equivariant (blue) and invariant (red) representations. Left: visualization of the pairwise Hamming distance along the single interpolation from Fig.~\ref{fig:mols_along_curve}.}
    \label{fig:hamming_dist}
\end{figure*}

Fig.~\ref{fig:mols_along_curve} compares the equivariant interpolation $s(\alpha z_i + (1- \alpha)z_j)$ with the isometric invariant interpolation $\alpha s(z_i) + (1-\alpha) s(z_j)$ by visualizing 25 decoded molecules sampled along each of the interpolating lines (left). The same two interpolating lines are compared through the lens of the isometric, invariant representation $\mathcal{Z}_s$ (right). We see that interpolation in the latent space yields pathological behavior, where molecule structure varies greatly along the line, whereas interpolation using the isometric, invariant cross section yields a far more smooth variation along the curve.

\begin{figure*}[htbp!]
    \centering
    \includegraphics[width=\textwidth]{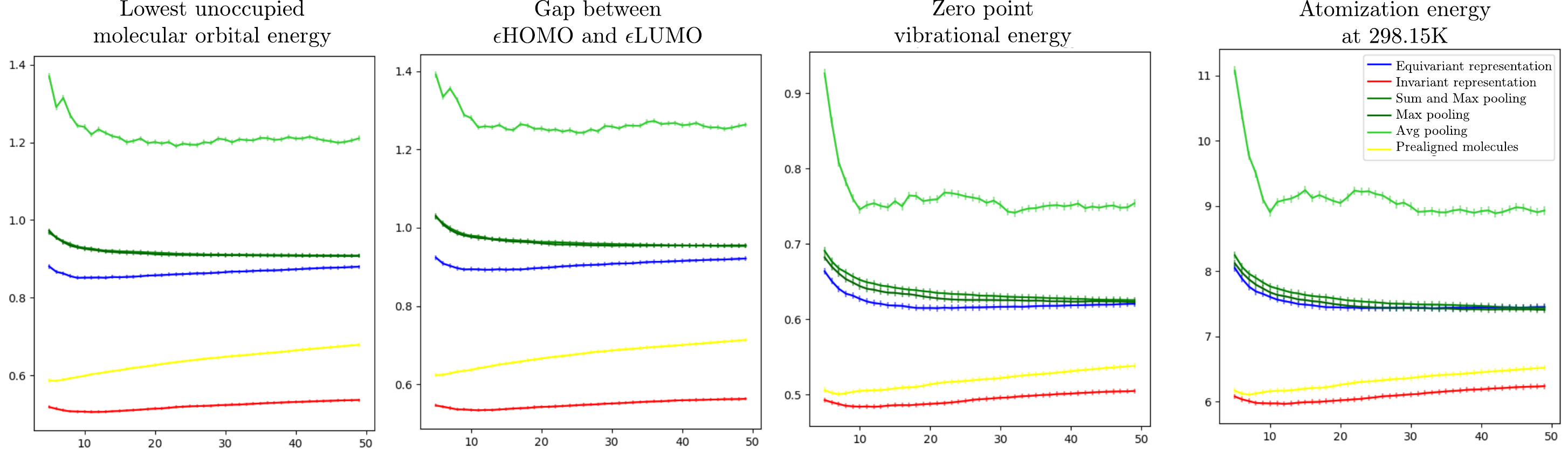}
    \caption{MAE vs $k$ for a kNN regression model predicting molecular properties from latent representations. We show MAE for the equivariant (blue) and isometric invariant representation (red), as well as invariant pooling operations commonly used to summarize equivariant representations.}
    \label{fig:within_neighbourhood_variance}
\end{figure*}

We confirm this tendency of difference in interpolation smoothness by computing the pairwise Hamming distance between consecutive molecules sampled along the interpolating line. Fig.~\ref{fig:hamming_dist} shows quantitatively that consecutive graphs generated by interpolating in the equivariant representation are less stable than those generated by interpolating in the invariant, isometric representation.

\paragraph{Stability of Molecules in a Neighborhood.}
Using the test-set containing approximately 13.000 molecules, that is, 10\% of the available data, we now compare the local structure of the equivariant and invariant representations using a $k$-nearest neighbor regression model to predict molecular properties from latent representations. We report the mean absolute error (MAE), as is custom for QM9 \cite{Wu2018-oo}, as a function of the number of neighbors considered, for each of the properties in Figure \ref{fig:within_neighbourhood_variance}.  In addition to the isometric invariant representation, we also consider several invariant pooling operations commonly used to summarize equivariant representations. It is clear that the MAE is consistently lowest when training a predictor on the isometric invariant representations, indicating that they are better at preserving information needed for downstream analysis.

\subsection{Rotation Invariant MNIST Classifier}
\begin{figure}
    \centering
        \includegraphics[width=0.33\textwidth]{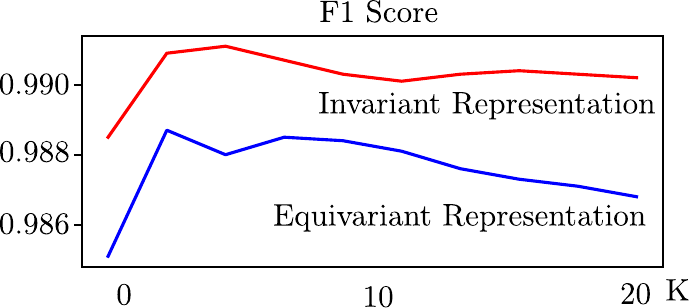}
    \caption{MNIST dataset classification problem: F1 score of KNN classifier applied to Invariant and Equivariant representations of the latent space.}
    \label{fig:f1_score}
\end{figure}
We demonstrate random invariant projections of equivariant latent features for a rotation invariant classifier trained on MNIST~\cite{lecun-mnisthandwrittendigit-2010} augmented by rotation. Let $h \colon \mathcal{X} \xrightarrow{f} \mathcal{Z} \xrightarrow{k} \mathcal{Y}$ be a model where $\mathcal{X}$ is the space of images acted on by the rotation group $G = SO(2)$. Let $f$ be an equivariant feature embedding derived from the $E(n)$-equivariant architecture of~\citep{Cesa2022-to}, followed by an invariant pooling layer $k$ and a fully connected classifier, giving a rotation invariant classifier see Appendix \ref{app:architecture-classifier} for a description of the full architecture.

\begin{figure*}
    \centering
    \includegraphics[angle=270,width=\textwidth, trim=6cm 0cm 6cm 0cm]{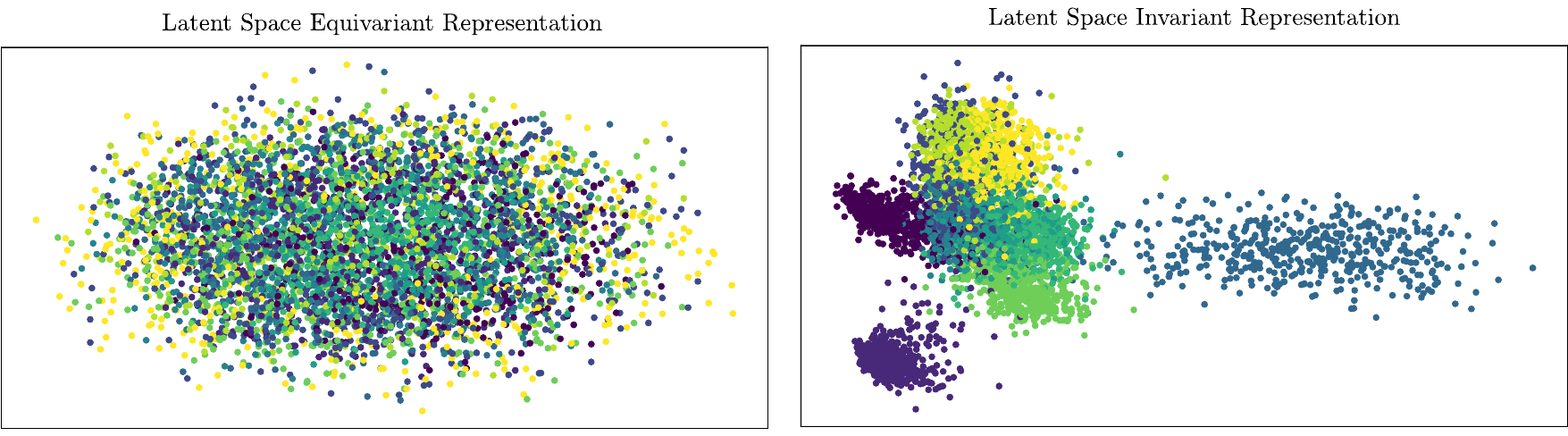}
    \caption{The figure shows the first two principal components of the MNIST training data for the equivariant representation (right), and the invariant representation (left) of the latent space. The colours represent the classifier labels.}
    \label{fig:pca_rotation}
\end{figure*}

The first two principal components of the equivariant and invariant representations are shown in Fig.~\ref{fig:pca_rotation}. We see that naively using the equivariant representation yields a plot suggesting very little signal. However, after applying a random linear projection, the structure becomes strikingly clear. We analyze the local structure of the representations by applying a $k$-nearest neighbor classifier. Its quality is evaluated using the F1 score; see Fig.~\ref{fig:f1_score}. Again, we see that the quality of the classifier increases when applied to the invariant representation as opposed to the equivariant one.

\section{Discussion}

We have shown, for two commonly encountered types of transformations, how equivariant latent representations can lead to inappropriate and ambiguous interpretations, as they contain multiple representations per data point. Moreover, we have explained how equivariant latent representations implicitly encode well-defined quotient representations. We show that for a particular permutation equivariant representation, this quotient representation $\mathcal{Z}$ admits an isometric cross section onto a well-defined subset $\mathcal{Z}_s$, retaining all information from the quotient $\mathcal{Z}/G$. More generally, we show how random invariant projections can produce informative invariant representations.

\paragraph{Is the need for post hoc analysis a problem?} Post hoc methods are sometimes considered inferior to intrinsic methods. We emphasize that the invariant representations presented here are comparable to dimensionality reduction methods such as UMAP, T-SNE or PCA -- which are also post hoc. The strength of our invariant representations is that they do not restrict the equivariant models themselves. This allows training models with performance only in mind, still obtaining invariant representations that are functions of the equivariant representations learned in model training.

\begin{figure}
    \centering
    \includegraphics[width=0.6\linewidth]{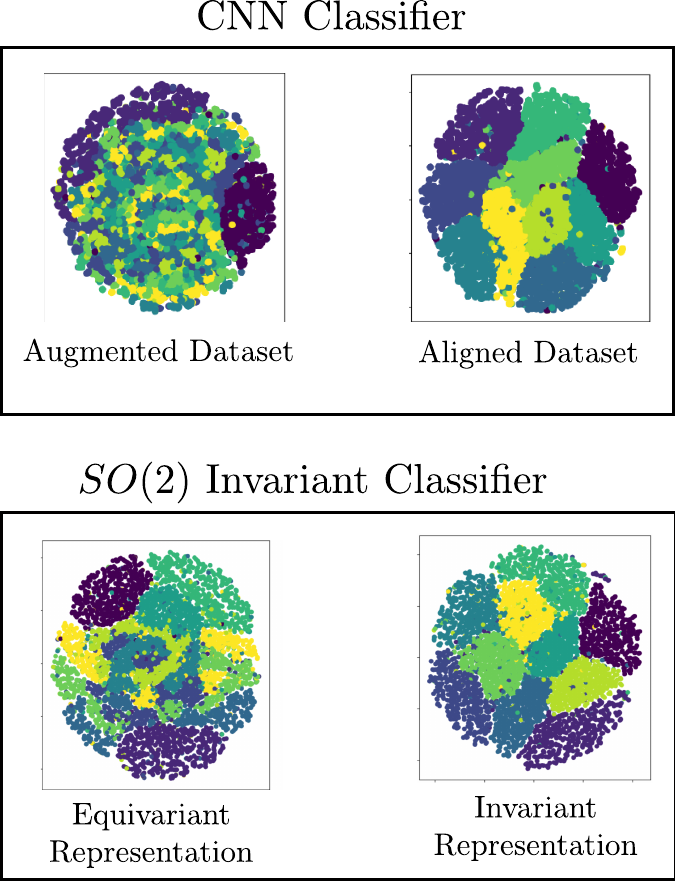}
    \caption{t-SNE plots showing the demonstrating how image alignment given by the natural orientation of digits provides a poor man's analogue of the invariant representation for the $SO(2)$ invariant classifier
    }
    \label{fig:augmentation_visualization}
\end{figure}

\paragraph{Weaknesses \& future work:} While choosing an invariant mapping is a natural extension of the assumption of equivariant modeling, it may not always be clear how this map should be chosen, as this choice may depend greatly on the model architecture. Suggesting suitable high-quality invariant mappings relevant for widely applied equivariant architectures is an obvious path for future work.

\paragraph{Isn't this just an argument to avoid the mathematically cumbersome equivariant and invariant models?} Invariant and equivariant models rely on advanced mathematical machinery that makes them less accessible than standard deep learning models. One might be tempted to take our results as an argument to just avoid these models altogether. However, this would not suffice to avoid our illustrated problems. An alternative and common way to encourage invariance or equivariance is to rely on the inherent flexibility of deep learning models to learn approximate invariance/equivariance through data augmentation. Fig.~\ref{fig:augmentation_visualization} illustrates the effect of using augmentation to encourage rotation invariance in a classification CNN trained on MNIST~\cite{deng2012mnist} -- we see a very similar pattern as in the previously described $SO(2)$ invariant classifier (bottom). On the top plot, we see how an intermediate latent representation of the CNN represents the MNIST test images when they are rotated randomly ("Augmented Dataset", left) or embedded using their original orientation ("Aligned Dataset", right). Since MNIST images have an orientation, they are naturally aligned with each other, and the resulting t-SNE plot nicely separates the different digits, obtaining an effect similar to the $SO(2)$ invariant classifier. For most image classification problems, however, there is no natural alignment, and a more realistic scenario is the plot on the left, where random rotations have been applied prior to embedding. What this shows is that while a pre-embedding alignment can work as a poor man's invariant representation, the lack of well-defined relative distances that we saw for strictly equivariant representations cannot be avoided by sticking to more straightforward models.

\section*{Acknowledgements}
This work was supported in part by the Independent Research Fund Denmark (grant no. 1032-00349B), by the Novo Nordisk Foundation through the Center for Basic Machine Learning Research in Life Science (grant no. NNF20OC0062606), the Pioneer Centre for AI, DNRF grant nr P1, and by the ERC Advanced grant $786854$ on Geometric Statistics.

\section*{Impact Statement}
This paper presents work whose goal is to advance the field of 
Machine Learning. There are many potential societal consequences 
of our work, none which we feel must be specifically highlighted here.


\bibliography{bibs}

\begin{thebibliography}{51}
\providecommand{\natexlab}[1]{#1}
\providecommand{\url}[1]{\texttt{#1}}
\expandafter\ifx\csname urlstyle\endcsname\relax
  \providecommand{\doi}[1]{doi: #1}\else
  \providecommand{\doi}{doi: \begingroup \urlstyle{rm}\Url}\fi

\bibitem[Aslan et~al.(2023)Aslan, Platt, and Sheard]{aslan2023group}
Aslan, B., Platt, D., and Sheard, D.
\newblock Group invariant machine learning by fundamental domain projections.
\newblock In Sanborn, S., Shewmake, C., Azeglio, S., Di~Bernardo, A., and Miolane, N. (eds.), \emph{Proceedings of the 1st {NeurIPS} Workshop on Symmetry and Geometry in Neural Representations}, volume 197 of \emph{Proceedings of Machine Learning Research}, pp.\  181--218. PMLR, December 2023.

\bibitem[Berthelot et~al.(2018)Berthelot, Raffel, Roy, and Goodfellow]{Berthelot2018-hh}
Berthelot, D., Raffel, C., Roy, A., and Goodfellow, I.
\newblock Understanding and improving interpolation in autoencoders via an adversarial regularizer.
\newblock July 2018.

\bibitem[Bredon(1972)]{bredon1972introduction}
Bredon, G.~E.
\newblock \emph{Introduction to compact transformation groups}.
\newblock Academic press, 1972.

\bibitem[Bronstein et~al.(2021)Bronstein, Bruna, Cohen, and Veli{\v c}kovi{\'c}]{Bronstein2021-fi}
Bronstein, M.~M., Bruna, J., Cohen, T., and Veli{\v c}kovi{\'c}, P.
\newblock Geometric deep learning: Grids, groups, graphs, geodesics, and gauges.
\newblock April 2021.

\bibitem[Calissano et~al.(2024)Calissano, Feragen, and Vantini]{calissano2024populations}
Calissano, A., Feragen, A., and Vantini, S.
\newblock Populations of unlabelled networks: Graph space geometry and generalized geodesic principal components.
\newblock \emph{Biometrika}, 111\penalty0 (1):\penalty0 147--170, 2024.

\bibitem[Candes \& Tao(2006)Candes and Tao]{candes2006near}
Candes, E.~J. and Tao, T.
\newblock Near-optimal signal recovery from random projections: Universal encoding strategies?
\newblock \emph{IEEE transactions on information theory}, 52\penalty0 (12):\penalty0 5406--5425, 2006.

\bibitem[Cesa et~al.(2022{\natexlab{a}})Cesa, Lang, and Weiler]{Cesa2022-to}
Cesa, G., Lang, L., and Weiler, M.
\newblock A program to build {E(N)}-equivariant steerable {CNN}s.
\newblock In \emph{International Conference on Learning Representations}, 2022{\natexlab{a}}.
\newblock URL \url{https://openreview.net/forum?id=WE4qe9xlnQw}.

\bibitem[Cesa et~al.(2022{\natexlab{b}})Cesa, Lang, and Weiler]{cesa2022a}
Cesa, G., Lang, L., and Weiler, M.
\newblock A program to build {E(N)}-equivariant steerable {CNN}s.
\newblock In \emph{International Conference on Learning Representations}, 2022{\natexlab{b}}.
\newblock URL \url{https://openreview.net/forum?id=WE4qe9xlnQw}.

\bibitem[Cohen \& Welling(2016)Cohen and Welling]{pmlr-v48-cohenc16}
Cohen, T. and Welling, M.
\newblock Group equivariant convolutional networks.
\newblock In Balcan, M.~F. and Weinberger, K.~Q. (eds.), \emph{Proceedings of The 33rd International Conference on Machine Learning}, volume~48 of \emph{Proceedings of Machine Learning Research}, pp.\  2990--2999, New York, New York, USA, 20--22 Jun 2016. PMLR.
\newblock URL \url{https://proceedings.mlr.press/v48/cohenc16.html}.

\bibitem[Deng(2012)]{deng2012mnist}
Deng, L.
\newblock The {MNIST} database of handwritten digit images for machine learning research [best of the web].
\newblock \emph{IEEE signal processing magazine}, 29\penalty0 (6):\penalty0 141--142, 2012.

\bibitem[Detlefsen et~al.(2022)Detlefsen, Hauberg, and Boomsma]{detlefsen2022learning}
Detlefsen, N.~S., Hauberg, S., and Boomsma, W.
\newblock Learning meaningful representations of protein sequences.
\newblock \emph{Nature communications}, 13\penalty0 (1):\penalty0 1914, 2022.

\bibitem[Feragen \& Nye(2020)Feragen and Nye]{feragen2020statistics}
Feragen, A. and Nye, T.
\newblock Statistics on stratified spaces.
\newblock In \emph{Riemannian geometric statistics in medical image analysis}, pp.\  299--342. Elsevier, 2020.

\bibitem[Fey \& Lenssen(2019)Fey and Lenssen]{Fey/Lenssen/2019}
Fey, M. and Lenssen, J.~E.
\newblock Fast graph representation learning with {PyTorch Geometric}.
\newblock In \emph{ICLR Workshop on Representation Learning on Graphs and Manifolds}, 2019.

\bibitem[Goodfellow et~al.(2020)Goodfellow, Pouget-Abadie, Mirza, Xu, Warde-Farley, Ozair, Courville, and Bengio]{goodfellow2020generative}
Goodfellow, I., Pouget-Abadie, J., Mirza, M., Xu, B., Warde-Farley, D., Ozair, S., Courville, A., and Bengio, Y.
\newblock Generative adversarial networks.
\newblock \emph{Communications of the ACM}, 63\penalty0 (11):\penalty0 139--144, 2020.

\bibitem[He et~al.(2022)He, Chen, Xie, Li, Doll{\'a}r, and Girshick]{he2022masked}
He, K., Chen, X., Xie, S., Li, Y., Doll{\'a}r, P., and Girshick, R.
\newblock Masked autoencoders are scalable vision learners.
\newblock In \emph{Proceedings of the IEEE/CVF conference on computer vision and pattern recognition}, pp.\  16000--16009, 2022.

\bibitem[Hy \& Kondor(2023)Hy and Kondor]{hy2023multiresolution}
Hy, T.~S. and Kondor, R.
\newblock Multiresolution equivariant graph variational autoencoder.
\newblock \emph{Machine Learning: Science and Technology}, 4\penalty0 (1):\penalty0 015031, 2023.

\bibitem[Kingma et~al.(2019)Kingma, Welling, et~al.]{kingma2019introduction}
Kingma, D.~P., Welling, M., et~al.
\newblock An introduction to variational autoencoders.
\newblock \emph{Foundations and Trends{\textregistered} in Machine Learning}, 12\penalty0 (4):\penalty0 307--392, 2019.

\bibitem[Kirillov et~al.(2023)Kirillov, Mintun, Ravi, Mao, Rolland, Gustafson, Xiao, Whitehead, Berg, Lo, et~al.]{kirillov2023segment}
Kirillov, A., Mintun, E., Ravi, N., Mao, H., Rolland, C., Gustafson, L., Xiao, T., Whitehead, S., Berg, A.~C., Lo, W.-Y., et~al.
\newblock Segment anything.
\newblock \emph{arXiv preprint arXiv:2304.02643}, 2023.

\bibitem[Kolaczyk et~al.(2020)Kolaczyk, Lin, Rosenberg, Walters, and Xu]{kolaczyk2020averages}
Kolaczyk, E.~D., Lin, L., Rosenberg, S., Walters, J., and Xu, J.
\newblock Averages of unlabeled networks: geometric characterization and asymptotic behaviour.
\newblock \emph{The Annals of Statistics}, 48\penalty0 (1):\penalty0 514--538, 2020.

\bibitem[Kwon et~al.(2023)Kwon, Choi, and Ryu]{Kwon2023-gv}
Kwon, S., Choi, J.~Y., and Ryu, E.~K.
\newblock Rotation and translation invariant representation learning with implicit neural representations.
\newblock In \emph{Proceedings of the 40th International Conference on Machine Learning}, volume 202 of \emph{ICML'23}, pp.\  18037--18056. JMLR.org, July 2023.

\bibitem[LeCun \& Cortes(2010)LeCun and Cortes]{lecun-mnisthandwrittendigit-2010}
LeCun, Y. and Cortes, C.
\newblock {MNIST} handwritten digit database.
\newblock http://yann.lecun.com/exdb/mnist/, 2010.
\newblock URL \url{http://yann.lecun.com/exdb/mnist/}.

\bibitem[Liu et~al.(2018)Liu, Allamanis, Brockschmidt, and Gaunt]{Liu2018-rj}
Liu, Q., Allamanis, M., Brockschmidt, M., and Gaunt, A.
\newblock Constrained graph variational autoencoders for molecule design.
\newblock \emph{Adv. Neural Inf. Process. Syst.}, 31, 2018.

\bibitem[Ma et~al.(2018)Ma, Chen, and Xiao]{Ma2018-tq}
Ma, T., Chen, J., and Xiao, C.
\newblock Constrained generation of semantically valid graphs via regularizing variational autoencoders.
\newblock \emph{arxiv:1809.02630}, September 2018.

\bibitem[Mardia \& Dryden(1989)Mardia and Dryden]{mardia1989statistical}
Mardia, K. and Dryden, I.
\newblock The statistical analysis of shape data.
\newblock \emph{Biometrika}, 76\penalty0 (2):\penalty0 271--281, 1989.

\bibitem[Maron et~al.(2018{\natexlab{a}})Maron, Ben-Hamu, Shamir, and Lipman]{Maron2018-ol}
Maron, H., Ben-Hamu, H., Shamir, N., and Lipman, Y.
\newblock Invariant and equivariant graph networks.
\newblock \emph{arxiv:1812.09902}, December 2018{\natexlab{a}}.

\bibitem[Maron et~al.(2018{\natexlab{b}})Maron, Ben-Hamu, Shamir, and Lipman]{maron2018invariant}
Maron, H., Ben-Hamu, H., Shamir, N., and Lipman, Y.
\newblock Invariant and equivariant graph networks.
\newblock \emph{arxiv:1812.09902}, December 2018{\natexlab{b}}.

\bibitem[Maron et~al.(2019)Maron, Ben-Hamu, Serviansky, and Lipman]{Maron2019-zr}
Maron, H., Ben-Hamu, H., Serviansky, H., and Lipman, Y.
\newblock Provably powerful graph networks.
\newblock \emph{arxiv:1905.11136}, May 2019.

\bibitem[Mathieu et~al.(2019)Mathieu, Rainforth, Siddharth, and Teh]{Mathieu2019-pi}
Mathieu, E., Rainforth, T., Siddharth, N., and Teh, Y.~W.
\newblock Disentangling disentanglement in variational autoencoders.
\newblock In Chaudhuri, K. and Salakhutdinov, R. (eds.), \emph{Proceedings of the 36th International Conference on Machine Learning}, volume~97 of \emph{Proceedings of Machine Learning Research}, pp.\  4402--4412. PMLR, 2019.

\bibitem[McInnes et~al.(2018)McInnes, Healy, and Melville]{mcinnes2018umap}
McInnes, L., Healy, J., and Melville, J.
\newblock Umap: Uniform manifold approximation and projection for dimension reduction.
\newblock \emph{arXiv:1802.03426}, 2018.

\bibitem[Mehr et~al.(2018)Mehr, Lieutier, Bermudez, Guitteny, Thome, and Cord]{mehr2018manifold}
Mehr, E., Lieutier, A., Bermudez, F.~S., Guitteny, V., Thome, N., and Cord, M.
\newblock Manifold learning in quotient spaces.
\newblock In \emph{Proceedings of the IEEE Conference on Computer Vision and Pattern Recognition}, pp.\  9165--9174, 2018.

\bibitem[Mitton et~al.(2021)Mitton, Senn, Wynne, and Murray-Smith]{Mitton2021-mz}
Mitton, J., Senn, H.~M., Wynne, K., and Murray-Smith, R.
\newblock A graph {VAE} and graph transformer approach to generating molecular graphs.
\newblock \emph{arxiv:2104.04345}, April 2021.

\bibitem[Munkres(2000)]{munkres2000topology}
Munkres, J.
\newblock \emph{Topology}.
\newblock Featured Titles for Topology. Prentice Hall, Incorporated, 2000.
\newblock ISBN 9780131816299.
\newblock URL \url{https://books.google.dk/books?id=XjoZAQAAIAAJ}.

\bibitem[Pan \& Kondor(2022)Pan and Kondor]{Pan2022-sg}
Pan, H. and Kondor, R.
\newblock Permutation equivariant layers for higher order interactions.
\newblock In Camps-Valls, G., Ruiz, F. J.~R., and Valera, I. (eds.), \emph{Proceedings of The 25th International Conference on Artificial Intelligence and Statistics}, volume 151 of \emph{Proceedings of Machine Learning Research}, pp.\  5987--6001. PMLR, 2022.

\bibitem[Papernot \& McDaniel(2018)Papernot and McDaniel]{papernot2018deep}
Papernot, N. and McDaniel, P.
\newblock Deep k-nearest neighbors: Towards confident, interpretable and robust deep learning.
\newblock \emph{arXiv:1803.04765}, 2018.

\bibitem[Paszke et~al.(2017)Paszke, Gross, Chintala, Chanan, Yang, DeVito, Lin, Desmaison, Antiga, and Lerer]{paszke2017automatic}
Paszke, A., Gross, S., Chintala, S., Chanan, G., Yang, E., DeVito, Z., Lin, Z., Desmaison, A., Antiga, L., and Lerer, A.
\newblock Automatic differentiation in pytorch.
\newblock In \emph{NIPS-W}, 2017.

\bibitem[Puny et~al.(2022)Puny, Atzmon, Smith, Misra, Grover, Ben-Hamu, and Lipman]{puny2022frame}
Puny, O., Atzmon, M., Smith, E.~J., Misra, I., Grover, A., Ben-Hamu, H., and Lipman, Y.
\newblock Frame averaging for invariant and equivariant network design.
\newblock In \emph{International Conference on Learning Representations}, 2022.
\newblock URL \url{https://openreview.net/forum?id=zIUyj55nXR}.

\bibitem[Ramakrishnan et~al.(2014)Ramakrishnan, Dral, Rupp, and von Lilienfeld]{Ramakrishnan2014-de}
Ramakrishnan, R., Dral, P.~O., Rupp, M., and von Lilienfeld, O.~A.
\newblock Quantum chemistry structures and properties of 134 kilo molecules.
\newblock \emph{Sci Data}, 1:\penalty0 140022, August 2014.

\bibitem[Rigoni et~al.(2020)Rigoni, Navarin, and Sperduti]{Rigoni2020-hg}
Rigoni, D., Navarin, N., and Sperduti, A.
\newblock Conditional constrained graph variational autoencoders for molecule design.
\newblock \emph{2020 IEEE Symposium Series on Computational Intelligence (SSCI)}, 2020.

\bibitem[Ruddigkeit et~al.(2012)Ruddigkeit, van Deursen, Blum, and Reymond]{Ruddigkeit2012-kh}
Ruddigkeit, L., van Deursen, R., Blum, L.~C., and Reymond, J.-L.
\newblock Enumeration of 166 billion organic small molecules in the chemical universe database {GDB-17}.
\newblock \emph{J. Chem. Inf. Model.}, 52\penalty0 (11):\penalty0 2864--2875, November 2012.

\bibitem[Sannai et~al.(2021)Sannai, Imaizumi, and Kawano]{sannai2021improved}
Sannai, A., Imaizumi, M., and Kawano, M.
\newblock Improved generalization bounds of group invariant / equivariant deep networks via quotient feature spaces, 2021.

\bibitem[Severn et~al.(2022)Severn, Dryden, and Preston]{severn2022manifold}
Severn, K.~E., Dryden, I.~L., and Preston, S.~P.
\newblock Manifold valued data analysis of samples of networks, with applications in corpus linguistics.
\newblock \emph{The Annals of Applied Statistics}, 16\penalty0 (1):\penalty0 368--390, 2022.

\bibitem[Simonovsky \& Komodakis(2018)Simonovsky and Komodakis]{simonovsky2018graphvae}
Simonovsky, M. and Komodakis, N.
\newblock Graphvae: Towards generation of small graphs using variational autoencoders.
\newblock In \emph{Artificial Neural Networks and Machine Learning--ICANN 2018: 27th International Conference on Artificial Neural Networks, Rhodes, Greece, October 4-7, 2018, Proceedings, Part I 27}, pp.\  412--422. Springer, 2018.

\bibitem[Thiede et~al.(2020)Thiede, Hy, and Kondor]{Thiede2020-sf}
Thiede, E.~H., Hy, T., and Kondor, R.
\newblock The general theory of permutation equivarant neural networks and higher order graph variational encoders.
\newblock \emph{CoRR}, abs/2004.03990, 2020.
\newblock URL \url{https://arxiv.org/abs/2004.03990}.

\bibitem[Van~der Maaten \& Hinton(2008)Van~der Maaten and Hinton]{tsne}
Van~der Maaten, L. and Hinton, G.
\newblock Visualizing data using {t-SNE}.
\newblock \emph{Journal of machine learning research}, 9\penalty0 (11), 2008.

\bibitem[Vignac \& Frossard(2021)Vignac and Frossard]{Vignac2021-lp}
Vignac, C. and Frossard, P.
\newblock {Top-N}: Equivariant set and graph generation without exchangeability.
\newblock \emph{arxiv:2110.02096}, October 2021.

\bibitem[Wang et~al.(2024)Wang, Hsu, Baker, Bertozzi, Xin, and Wang]{wang2024rethinking}
Wang, S.-H., Hsu, Y.-C., Baker, J., Bertozzi, A.~L., Xin, J., and Wang, B.
\newblock Rethinking the benefits of steerable features in 3d equivariant graph neural networks.
\newblock In \emph{The Twelfth International Conference on Learning Representations}, 2024.
\newblock URL \url{https://openreview.net/forum?id=mGHJAyR8w0}.

\bibitem[Weiler \& Cesa(2019)Weiler and Cesa]{e2cnn}
Weiler, M. and Cesa, G.
\newblock {General E(2)-Equivariant Steerable CNNs}.
\newblock In \emph{Conference on Neural Information Processing Systems (NeurIPS)}, 2019.

\bibitem[Williams et~al.(2021)Williams, Kunz, Kornblith, and Linderman]{Williams2021-ow}
Williams, A.~H., Kunz, E., Kornblith, S., and Linderman, S.~W.
\newblock Generalized shape metrics on neural representations.
\newblock \emph{Adv. Neural Inf. Process. Syst.}, 34:\penalty0 4738--4750, December 2021.

\bibitem[Winter et~al.(2021)Winter, Noe, and Clevert]{Winter_undated-pz}
Winter, R., Noe, F., and Clevert, D.-A.
\newblock Permutation-invariant variational autoencoder for graph-level representation learning.
\newblock In Ranzato, M., Beygelzimer, A., Dauphin, Y., Liang, P., and Vaughan, J.~W. (eds.), \emph{Advances in Neural Information Processing Systems}, volume~34, pp.\  9559--9573. Curran Associates, Inc., 2021.
\newblock URL \url{https://proceedings.neurips.cc/paper_files/paper/2021/file/4f3d7d38d24b740c95da2b03dc3a2333-Paper.pdf}.

\bibitem[Winter et~al.(2022)Winter, Bertolini, Le, No{\'e}, and Clevert]{Winter2022-pl}
Winter, R., Bertolini, M., Le, T., No{\'e}, F., and Clevert, D.-A.
\newblock Unsupervised learning of group invariant and equivariant representations.
\newblock \emph{arxiv:2202.07559}, February 2022.

\bibitem[Wu et~al.(2018)Wu, Ramsundar, Feinberg, Gomes, Geniesse, Pappu, Leswing, and Pande]{Wu2018-oo}
Wu, Z., Ramsundar, B., Feinberg, E.~N., Gomes, J., Geniesse, C., Pappu, A.~S., Leswing, K., and Pande, V.
\newblock {MoleculeNet}: a benchmark for molecular machine learning.
\newblock \emph{Chem. Sci.}, 9\penalty0 (2):\penalty0 513--530, January 2018.

\end{thebibliography}
\bibliographystyle{icml2024}

\newpage
\appendix
\onecolumn
\section{Proofs}
\label{app:proofs}
This section will contain the proofs of central statements used in the paper.

\subsection{Proof of Proposition \ref{prop:invariant_surjective}}
\label{proof:inv_sur}

First we show that $s'$ is well defined. Pick $z_1, z_2 \in \mathcal{Z}$ and assume $[z_1] = [z_2]$. Then, by definition, $z_1 \in [z_2]$, and as a consequence, there exists some permutation $\sigma \in S_n$ such that $z_1 = \sigma z_2$. We now see that $s'([z_1]) = s(z_1) = s(\sigma z_2) = s(z_2) = s'([z_2])$, where the third equality follows from $s$ being invariant, and thus $s'$ is well defined.

We now show, that $s'$ is indeed surjective. Pick $y \in \mathcal{Z}_s$. Since $s$ is surjective, we have that there exists $z \in \mathcal{Z}$ such that $s(z) = y$. However, then $s'([z])=s(z) = y$. Thus $s'$ is surjective.

\subsection{Proof of Proposition \ref{prop:cont-quotient-homeo}}
\label{proof:cont-quotient-homeo}
The following proof is heavily inspired by the proof of Theorem 22.2. and Corollary 22.3 from \cite{munkres2000topology}, but is included for the sake of completeness.

We first show that $s'$ is continuous if and only if $s$ is continuous. First assume $s$ is continuous. As the canonical projection $\pi: \mathcal{Z} \to \mathcal{Z}/G$ is a quotient map, hence continuous, the composition $s' \circ \pi = s$ is continuous. Now suppose $s$ is continuous. Then given an open set $U \subseteq \mathcal{Z}_s$ we have that $s^{-1}(U)$ is open in $\mathcal{Z}$. But since the diagram of proposition \ref{prop:invariant_surjective} commutes, we have $s^{-1}(U) = (s' \circ \pi)^{-1}(U)$. Since $\pi$ is a quotient it now follows that $s'^{-1}(U)$ is open in $\mathcal{Z}/G$.

Next, we show, that $s'$ is a quotient map if and only if $s$ is a quotient map. Assume $s'$ is a quotient map. Again, the composition $s' \circ \pi = s$ is a quotient map, as the composition of two quotient maps is itself a quotient map. Now suppose that $s$ is a quotient map. Since $s$ is surjective, we have that $s' \circ \pi = s$ is surjective, and hence $s'$ is surjective. Now consider a subset $U \subseteq \mathcal{Z}_s$, and assume $s'^{-1}(U)$ is open in $\mathcal{Z}/G$. Then $\pi^{-1}(s'^{-1}(U))$ is open in $\mathcal{Z}$ as $\pi$ is continuous. But then $s^(-1)(U) = \pi^{-1}(s'^{-1}(U))$ is also open in $\mathcal{Z}$, and as such, $U$ is open in $\mathcal{Z}_s$ as $s$ is a quotient map. In combination with the fact that $s'$ is continuous if and only if $s$ is continuous the assertion follows.

Lastly the assertion that $s'$ is a homeomorphism if and only if $s'$ is a bijective quotient map follows directly from the definitions.

\subsection{Proof of Proposition \ref{prop:graph-bijective-iso}}
\label{proof:prop:graph-bijective-iso}
\paragraph{$s'$ is bijective:} As a consequence of proposition 1, the function $s'$ is well defined, and surjective. It remains to show that $s'$ is injective. Pick $[z_1],[z_2] \in \mathcal{Z}/S_n$ and assume $[z_1] \neq [z_2]$. Then $[z_1] \cap [z_2] = \emptyset$, and thus there exists no permutation $\sigma \in S_n$ such that $\sigma z_1 = z_2$. Assume for contradiction that $s'([z_1]) = s'([z_2])$. Then we have that:
\begin{equation}
    \sigma_{z_1}z_1 =  s(z_1) =  s'([z_1]) = s'([z_2]) = s(z_2) = \sigma_{z_2}z_2
\end{equation}
However, then $z_2 = (\sigma_{z_2}^{-1}\sigma_{z_1})z_1$. But as $S_n$ is a group $\pi_{z_2}^{-1}\pi_{z_1} \in S_n$, and thus we have a contradiction.

\paragraph{$s'$ is an isometry:}
We aim to show that $d([x], [y]) = d_{\mathcal{Z}_s}\left(s'([x], s'([y])\right)$. For all $[x], [y] \in \mathcal{Z}/S_n$, where $d$ denotes the quotient metric.

Pick $[x], [y] \in \mathcal{Z}/S_n$. Assume without loss of generality that $s(x) = x$ and $s(y) = y$, i.e. that the coordinates of $x$ and $y$ are already sorted. We can assume this as the coordinates of any representative can be sorted, thus achieving representatives with the properties we seek. 

It now suffices to show that for any $\sigma \in S_n$:
\begin{equation}
    d^2_{\mathcal{Z}_s}(x, y) = \sum_{i=1}^n (x_i - y_i)^2 \leq \sum_{i=1}^n (x_{\sigma(i)} -y_{i})^2 = d^2_{\mathcal{Z}_s}(\sigma x, y)
\end{equation}
If $\sigma$ is the identity permutation, then the above equation is trivially true. Also we note, that any permutation, can be written as a product of transpositions, thus if the above inequality is true for transpositions, then it will generalize to any permutation. Assume that $\sigma$ is a transposition, that is for $1 \leq j \leq k \leq n$ we have that $\sigma(j)=k$, $\sigma(k)=j$ and $\sigma(i)=i$ for $i \neq j,k$. Then
\begin{align}
    d^2_{\mathcal{Z}_s}(\sigma x, y) &= \sum_{i=1}^n (x_{\sigma(i)} - y_i)^2 \\
    &= (x_k - y_j)^2 + (x_j - y_k)^2 + \sum_{i \in \{ i \mid \pi(i)=i\}}^n (x_{i} - y_i)^2 ,
\end{align}
and therefore it now suffices to show that when $x_j \leq x_k$ and $y_j \leq y_k$ then
\begin{equation}
     (x_j-y_j)^2 + (x_k-y_k)^2 \leq (x_k - y_j)^2 + (x_j - y_k)^2
\end{equation}
That this is indeed true, can be seen by observing that $x_k = x_j + c$ for some constant $c \geq 0$. Then 
\begin{align*}
    (x_j-y_j)^2 + (x_k - y_k)^2 &= x_j^2 + y_j^2 - 2 x_j y_j + x_k^2 + y_k^2 - 2 x_k y_k \\
    &= x_j^2 + y_j^2 - 2 (x_k - c) y_j + x_k^2 + y_k^2 - 2 (x_j + c) y_k \\
    &= (x_k - y_j)^2 + (x_j - y_k)^2 + 2c(y_j - y_k) \\
    &\leq (x_k - y_j)^2 + (x_j - y_k)^2
\end{align*}
where the last inequality follows as $2c(y_j - y_k) \leq 0$ since $c \geq 0$ and $y_j \leq y_k$ by assumption. The assertion follows.

\paragraph{$\mathcal{Z}_s$ is a convex cone:} We first show that $\mathcal{Z}_s$ is a cone, i.e. if $y \in \mathcal{Z}_s$ then $\alpha y \in \mathcal{Z}_s$ for all $\alpha \geq 0$. Assume $y \in \mathcal{Z}_s$. Then
\begin{equation}
    y_1 \leq y_2 \leq ... \leq y_n
\end{equation}
but then clearly
\begin{equation}
    \alpha y_1 \leq  \alpha y_2 \leq ... \leq  \alpha y_n
\end{equation}
which means that $\alpha y \in \mathcal{Z}_s$.
Now, we show that $\mathcal{Z}_s$ is convex. Assume $x, y \in \mathcal{Z}_s$. Then:
\begin{equation}
    x_1 \leq  x_2 \leq ... \leq  x_n \text{   and  } y_1 \leq  y_2 \leq ... \leq y_n
\end{equation}
But then 
\begin{equation}
    x_1 + y_1 \leq  x_2 + y_2 \leq ... \leq  x_n + y_n
\end{equation}
which implies that $x + y \in \mathcal{Z}_s$. Since $\mathcal{Z}_s$ is a cone it now follows that $\mathcal{Z}_s$ is also convex.

\section{Permutation equivariant VAE}
\label{app:architecture-vae}
 By \cite{Maron2018-ol, Maron2019-zr, Thiede2020-sf, Pan2022-sg} we have that we can define any linear function $L: \mathbb{R}^{n^k \times d} \to \mathbb{R}^{n^l \times d'}$ using exactly $b(k+l)dd'$ known basis elements, where $b(\cdot)$ denotes the Bell number. Using this result, we can define node- and edge-level linear equivariant layers:
\begin{align}
    L_V&: \mathbb{R}^{n \times d_v} \to \mathbb{R}^{n^2 \times d'_e} \\
    L_E&: \mathbb{R}^{n^2 \times d_e} \to \mathbb{R}^{n^2 \times d'_e} 
\end{align}
 by using the a weighted linear combination of the known basis elements. This amounts to a total of $8 d_v d'_v$ weights in the case of $L_V$ and $15 d_e d'_e$ weights in the case of $L_E$. For an exact construction $L_V$ and $L_E$ please refer to \cite{Pan2022-sg}. Using this construction, we can define a linear layer $L_{V,E}: \mathbb{R}^{n \times d_v} \times \mathbb{R}^{n^2 \times d_E} \to \mathbb{R}^{n^2 \times d'_E}$, by the channel-wise concatenation of $L(V)$ and $L(E)$. Note that this concatenation does not change the equivariance property of the layer. All linear layers of the architecture utilised in the current work is of one of these forms. Subsequently, a ReLU activation function is applied.

After each linear layer a 2D convolution with a $1 \times 1$ kernel size, a ReLU activation and instance normalization is applied. Again, all operations preserve the equivariance property of the network.

\paragraph{Encoder:} The encoder consists of a four linear layers:
\begin{itemize}
    \item The first layer is a \textit{hybrid} layer mapping the edge- and node-representation to a matrix representation similar to $L_{V,E}$.
    \item The two subsequent layers are equivariant linear layers mapping between matrix representations similar to the layers defined as $L_E$.
    \item The last layer, mapping to the latent representation, maps a matrix representation to a feature vector representation. In the current work we choose the number of latent channels to be $1$.
\end{itemize}
Each of the linear layers, except for the last layer, are followed by 2D convolutions, ReLU activations, and instance normalization as described above.

\paragraph{Decoder: } The decoder is likewise constructed using four linear layers:
\begin{itemize}
    \item The first layer maps the latent feature vector to a matrix representation similar to $L_V$.
    \item The two subsequent layers map are equivariant linear layers mapping between matrix representations.
    \item The last layer is then the concatenation of a linear layer mapping between matrix representations (the reconstructed edge-matrix) and a linear layer mapping to a feature vector representation (the reconstructed node-matrix). The reconstructed edgematrix is enforced to be symmetric by adding the transpose of itself.
\end{itemize}
Again, following each layer except the last, we apply ReLU activations and instance normalization. In the last layer, a pointwise softmax is applied.

\paragraph{Training details:} The model was trained using the negative evidence lower bound (ELBO) as is standard for VAEs. A learning rate of 0.0001 and a batch-size of 32 was chosen. The model was trained for 1000 epochs. The QM9 dataset was obtained through the Python Geometric library \cite{Fey/Lenssen/2019}.

\subsection{Model}
We assume the following likelihood model of the examined graphs:
\begin{equation}
    \log p_\theta(G|\mathbf{z}) = \log p_\theta(V|\mathbf{z}) + \log p_\theta(E|\mathbf{z})
\end{equation}
From the perspective of a variational autoencoder, $\theta$ denotes the parameters of the \textit{decoder}. We assume that each node vector $\mathbf{v} \in V$ and edge vector $\mathbf{c} \in E$ are obtained independently from a categorical distribution parameterized by $\theta$. That is:
\begin{align}
    p_\theta(V | \mathbf{z}) &= \prod_{\mathbf{v} \in V} p_\theta(\mathbf{v} | \mathbf{z}) \\
    p_\theta(E | \mathbf{z}) &= \prod_{\mathbf{c} \in E} p_\theta(\mathbf{c} | \mathbf{z}),
\end{align}
where $p_\theta(\mathbf{v} | \mathbf{z})$ and $p_\theta(\mathbf{c} | \mathbf{z})$ respectfully denotes the probablility of observing $\mathbf{v}$ and $\mathbf{c}$ respectfully under the model.

Following the standard notation for VAEs, we impose the prior $p(\mathbf{z}) = \mathcal{N}(\textbf{0} | \mathbf{1})$, and as a consequence we naturally let the approximate posterior be given as:
\begin{equation}
    q_\phi(\mathbf{z}|G) = \mathcal{N}(\mathbf{\mu}_\phi(G) | \mathbf{\sigma}^2_{\mathbf{\phi}}(G))
\end{equation}
where, $\phi$ denotes the parameters of the \textit{encoder} of the VAE. Our objective is to minimize the negative evidence lower bound (ELBO) of the log-likelihood $\log(p_\theta(G))$, i.e.:
\begin{align}
    -ELBO 
    &= \mathbb{E}_{\mathbf{z} \sim q_\phi(\mathbf{z} | G)}[- \log p_\theta(G | \mathbf{z}))] + KL(q_\phi(\mathbf{z} | G) || p_\theta(\mathbf{z})) \\
    &= \mathbb{E}_{\mathbf{z} \sim q_\phi(\mathbf{z}|G)}\left[-\log\left(\frac{p_\theta(G|\mathbf{z})p(\mathbf{z})}{q_\phi(\mathbf{z}|G)}\right)\right],
\end{align}
where $KL(\cdot||\cdot)$ denotes the Kullback Leibler (KL) divergence.

\section{Rotation invariant MNIST-classifier}
\label{app:architecture-classifier}
The SO(2) invariant MNIST classifier is constructed using the tools provided in the ESCNN library provided by \cite{e2cnn, cesa2022a}. That is, the model consists of 6 SO(2) steerable planar convolutions each of which is followed by batch-normalization and the FourierELU activation function. Each steerable planar convolution uses two irreducible representations to describe the output type; one invariant and one equivariant. The first two planar convolutions contains 16 feature maps, the next two 32 feature maps and the last two 64 feature maps. After each pair of planar convolutions, a pooling layer is applied, and after the last convolution pooling is done over the spatial dimension to ensure invariance. Lastly, an invariant classifier is appended to the model. This is implemented using specifying a convolution using a kernel size of $1 \times 1$, and using the trivial representation to describe the output type, followed by a fully-connected classification network.

\paragraph{Training details:} The model was trained using a cross-entropy loss. A learning rate of 0.01 and a batch-size of 128 was chosen. The model was trained for 100 epochs. The MNIST dataset was obtained through the pytorch library\cite{paszke2017automatic}.


\end{document}